\documentclass{article}

\usepackage{microtype}
\usepackage{graphicx}
\usepackage{subfigure}
\usepackage{booktabs} 
\usepackage{multirow}

\usepackage{colortbl}
\usepackage{hyperref}
\usepackage{wrapfig}

\usepackage[accepted]{icml2023}

\usepackage{amsmath}
\usepackage{amssymb}
\usepackage{mathtools}
\usepackage{amsthm}

\usepackage[capitalize,noabbrev]{cleveref}

\theoremstyle{plain}

\theoremstyle{definition}

\theoremstyle{remark}

\definecolor{ygreen}{rgb}{0.678,0.757,0.471}
\definecolor{ybrown}{rgb}{0.663, 0.518, 0.404}
\definecolor{ydarky}{rgb}{0.424, 0.345, 0.298}

\usepackage[textsize=tiny]{todonotes}

\icmltitlerunning{Robust Weight Signatures: Gaining Robustness as Easy as Patching Weights?}

\begin{document}

\twocolumn[
\icmltitle{Robust Weight Signatures: Gaining Robustness as Easy as Patching Weights?}




\begin{icmlauthorlist}
\icmlauthor{Ruisi Cai}{yyy}
\icmlauthor{Zhenyu Zhang}{yyy}
\icmlauthor{Zhangyang Wang}{yyy} \\
\vspace{0.3em}
{\textsuperscript{1}VITA Group, University of Texas at Austin}

\end{icmlauthorlist}

\icmlaffiliation{yyy}{Department of Electrical and Computer Engineering, University of Texas at Austin}

\icmlcorrespondingauthor{Zhangyang Wang}{atlaswang@utexas.edu}

\icmlkeywords{Machine Learning, ICML, model soup, Robustness, Weight Aggregation}

\vskip 0.15in
]




\begin{abstract}
\vspace{-0.5em}
Given a robust model trained to be resilient to one or multiple types of distribution shifts (e.g., natural image corruptions), \textit{how is that ``robustness'' encoded in the model weights, and how easily can it be disentangled and/or ``zero-shot" transferred to some other models}? This paper empirically suggests a surprisingly simple answer: \textbf{linearly - by straightforward model weight arithmetic}! We start by drawing several key observations: (i) assuming that we train the same model architecture on both a clean dataset and its corrupted version, a comparison between the two resultant models shows their weights to mostly differ in shallow layers; (ii) the weight difference after projection, which we call \textit{``Robust Weight Signature"} (\textbf{RWS}), appears to be discriminative and indicative of different corruption types; (iii) perhaps most strikingly, for the same corruption type, the RWSs obtained by one model architecture are highly consistent and transferable across different datasets. 

Based on those RWS observations, we propose a minimalistic model robustness ``patching" framework that carries a model trained on clean data together with its pre-extracted RWSs. In this way, injecting certain robustness to the model is reduced to directly adding the corresponding RWS to its weight. We experimentally verify our proposed framework to be remarkably (1) \textbf{lightweight}. since RWSs concentrate on the shallowest few layers and we further show they can be painlessly quantized, storing an RWS is up to 13 $\times$ more compact than storing the full weight copy; (2) \textbf{in-situ adjustable}. RWSs can be appended as needed and later taken off to restore the intact clean model. We further demonstrate one can linearly re-scale the RWS to control the patched robustness strength; (3) \textbf{composable}. Multiple RWSs can be added simultaneously to patch more comprehensive robustness at once; and (4) \textbf{transferable}. Even when the clean model backbone is continually adapted or updated, RWSs remain as effective patches due to their outstanding cross-dataset transferability. 

\vspace{-1em}
\end{abstract}

\section{Introduction}

\subsection{Background and Related Work}
\vspace{-0.2em}


The robustness and safety of machine learning models have become prevailing concerns for practitioners. Among many other possible forms of safety risks such as adversarial attacks \cite{madry2017towards,zhang2019theoretically} and backdoor attacks \cite{goldblum2022dataset}, one concern of particular significance is the model's resilience against various distribution shifts from training data \cite{koh2021wilds}. For example, a computer vision model for autonomous driving or video surveillance could be trained on relatively ``clean" and constrained data to achieve high performance on standard benchmarks. However, they are vulnerable to unforeseen distributional changes including natural corruptions (e.g., due to camera noise, motion blur, adverse weather), sensory perturbations (e.g., sensor transient error, electromagnetic interference), and larger domain shift forms (e.g., summer → winter, daytime → night) - hence jeopardizing their trustworthiness and safe deployment. 

Many solutions have since been examined to strengthen the models' robustness against unforeseen distribution shifts, in particular \underline{natural image corruptions} - which would be the focus of this paper. Examples include data augmentation \cite{hendrycks2021many,hendrycks2019augmix,rusak2020simple,wang2021augmax}, stability-aware training \cite{hein2017formal,zheng2016improving}, leveraging pre-trained models \cite{hendrycks2019using,chen2020adversarial,jiang2020robust,sun2021improving,wortsman2022robust} or training on larger and more diverse datasets \cite{taori2020measuring,nguyen2022quality}. Among them, data augmentation is perhaps the most popular practice, as it is easy to implement and plug in. It also remains as the most empirically effective approach to gain comprehensive robustness to various natural corruptions \cite{hendrycks2019augmix,wang2021augmax}, though at the cost of training time overhead.

Further complicating the problem is the inherent ``trade-off" between model standard accuracy and robustness, informally: \textit{the more ``comprehensive" robustness that a model strives to cover, the less ``focused" it can fit the standard clean data distribution}. Firstly observed in \cite{tsipras2019robustness}, the authors pointed out that adversarial training (AT) \cite{madry2018towards}, which utilizes adversarial samples as a special data augmentation method, has also shown to improve model robustness yet sacrificing the standard accuracy on clean images. The same trade-off observation holds generally true for other data augmentation and stability-aware training methods \cite{wang2021augmax}, essentially reflecting the ``bias-variance" trade-off. Practically, most defense methods determine their accuracy-robustness trade-off by some empirically hyper-parameter pre-chosen at training, such as the coefficient weight between the standard and robust classification losses for AT, or the strength of data augmentations. Such methods will hence ``pre-fix" achievable standard and robust accuracies at training time, leaving no flexibility to adjust for testing even if there is a demand.



In practical AI platforms especially at the edge, the desired trade-off between standard and robust accuracies often varies adaptively depending on contexts, which are not always met by the pre-fixed ``default” settings. For example, an autonomous agent might perceive using its ``standard" mode for normal-environment operations (most of the time), but switch to behaving more conservatively such as when placed in less familiar or adverse environments. Re-training the model, especially robustly, is notoriously resource-consuming and impossible for in-situ adjustment. Hence practitioners look for convenient means to explore and flexibly calibrate the accuracy-robustness trade-off at the testing time. Test-time adaption \cite{fleuret2021test,croce2022evaluating} or ensembling \cite{liu2018towards} methods, though effective, will turn impractical when memory and storage are in limited supply or the inference latency is sensitive. The recent ``once-for-all" AT methods \cite{wang2020once,kundu2023float} enable the network to adjust to different input distributions nearly free of overheads, by input-conditioning. However, all aforementioned methods would compromise the achievable clean accuracy more or less, in exchange for encapsulating more robustness in the same model. Also most of them focus on adversarial attacks.

\begin{figure}[tb]
    \centering
    \includegraphics[width=1.0\linewidth]{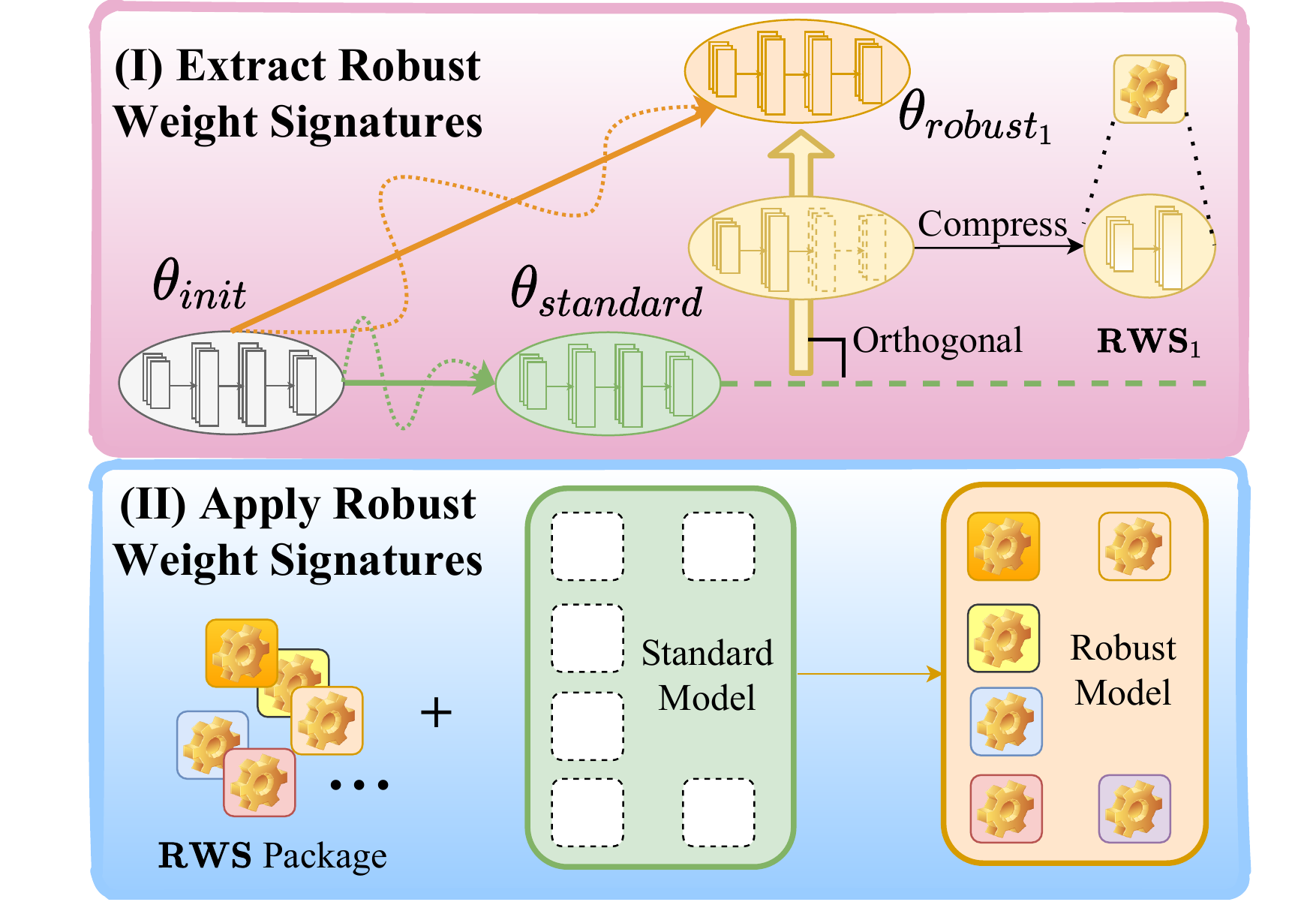}
    \vspace{-0.5em}
    \caption{Overview of our pipeline: Step (\uppercase\expandafter{\romannumeral1}): Extract Robust Weight Signatures (RWSs) by comparing the difference between robust models and standard models of shallow layers in the weight space. Step (\uppercase\expandafter{\romannumeral2}): Patch non-robust models by RWSs as needed.}
    \vspace{-1em}
    \label{fig:teasor}
\end{figure}
\subsection{Our Aim and Contributions}

This paper targets the ``in-situ" adaptive robustness similarly as defined in \cite{wang2020once,kundu2023float}, i.e., to painlessly calibrate on the accuracy-robustness trade-off at the test time, with minimal overhead in latency or memory. Our \textbf{problem setting and goal} will yet differ in (1) focusing on the comprehensive robustness against unforeseen natural image corruptions (\textbf{not} adversarial attacks); and (2) \textbf{not} impairing the standard accuracy on clean test images at all.

Our proposal is a minimalistic model robustness ``patching” framework that differs remarkably from the aforementioned efforts. We are inspired by the recent findings on the linear interpolatability between model weights (such as pre-trained and fine-tuned) \cite{wortsman2022model,wortsman2022robust,li2022branch,ilharco2022editing}. Contrary to the common wisdom of model output ensembling, \cite{wortsman2022model} pioneered averaging the weights of multiple fine-tuned models directly, without incurring any additional inference or memory costs, that yield significantly improved ``zero-shot" and out-of-distribution generalization performance. Most relevantly, \cite{ilharco2022editing} demonstrated that such model weight ``arithmetic" can go beyond averaging: by computing the weight difference between a pre-trained model and its downstream-task fine-tuned version (called a ``task vector"); the resulting task vectors are noted to meaningfully steer the behavior of neural networks: they can be modified and combined together through arithmetic operations such as negation and addition, e.g., adding multiple task vectors together can improve performance on multiple tasks at once. 

In view of those, we ask: in a robust model, \textit{how is that "robustness" encoded in the model weight space, and how can it be decoded, combined, or transferred? How would that further help our in-situ adaptive robustness goal?} 
Given a standard model (trained on clean data) and its robust counterpart (trained on corrupted versions of the same dataset), we compare their difference to extract the \textit{``Robust Weight Signature"} (\textbf{RWS}). 
It turns out that RWS lends a surprisingly effective, elegantly simple and flexible means to achieve in-situ robustness, due to the following novel findings:
\begin{itemize}
\vspace{-1em}
    \item Assuming a model trained on clean data together with its pre-extracted RWSs to multiple image corruption types, ``patching" certain robustness to the model is reduced to directly adding the corresponding RWS to its weight. Any appended RWS can be later taken off to switch back to the intact clean model: hence there is no compromise of standard accuracy.\vspace{-0.2em} 
    \item RWSs are highly compressible since their large-magnitude elements are dominantly in the lower layers. We further show them to be robust to quantization as well. Hence storing an RWS is up to 13$\times$ more compact than the full weight copy, mitigating the storage burden of carrying multiple pre-trained models.\vspace{-0.2em} 
    \item RWSs are extraordinarily controllable and combinable: one can linearly re-scale an RWS to control the patched robustness strength. Multiple RWSs can be added simultaneously to patch more comprehensive robustness at once. Essentially, we demonstrate the task ``arithmetic" claims in \cite{ilharco2022editing}  to be generally valid for multiple robustness types as well.\vspace{-0.2em}
    \item Lastly and uniquely, we find that an RWS is not tied with the standard model where it is subtracted. That is, when the standard model is updated, continually adapted, or even completely re-trained on a different dataset, the RWS seems to be the same applicable to the new model (same architecture). Such outstanding cross-data transferability decouples the standard model updating and robustness preservation, potentially saving training costs and boosting weight re-usability.\vspace{-0.5em}
\end{itemize}
In what follows, we accompany our claims with experimental results, showing that RWSs extensively improve model robustness to various natural image corruptions in a plug-and-play manner, while demonstrating to be \textbf{lightweight}, \textbf{in-situ adjustable}, \textbf{composable}, and \textbf{transferrable}. Our codes are available at: \url{https://github.com/VITA-Group/Robust_Weight_Signatures}.

\begin{figure}[tb]
    \centering
    \includegraphics[width=0.9\linewidth]{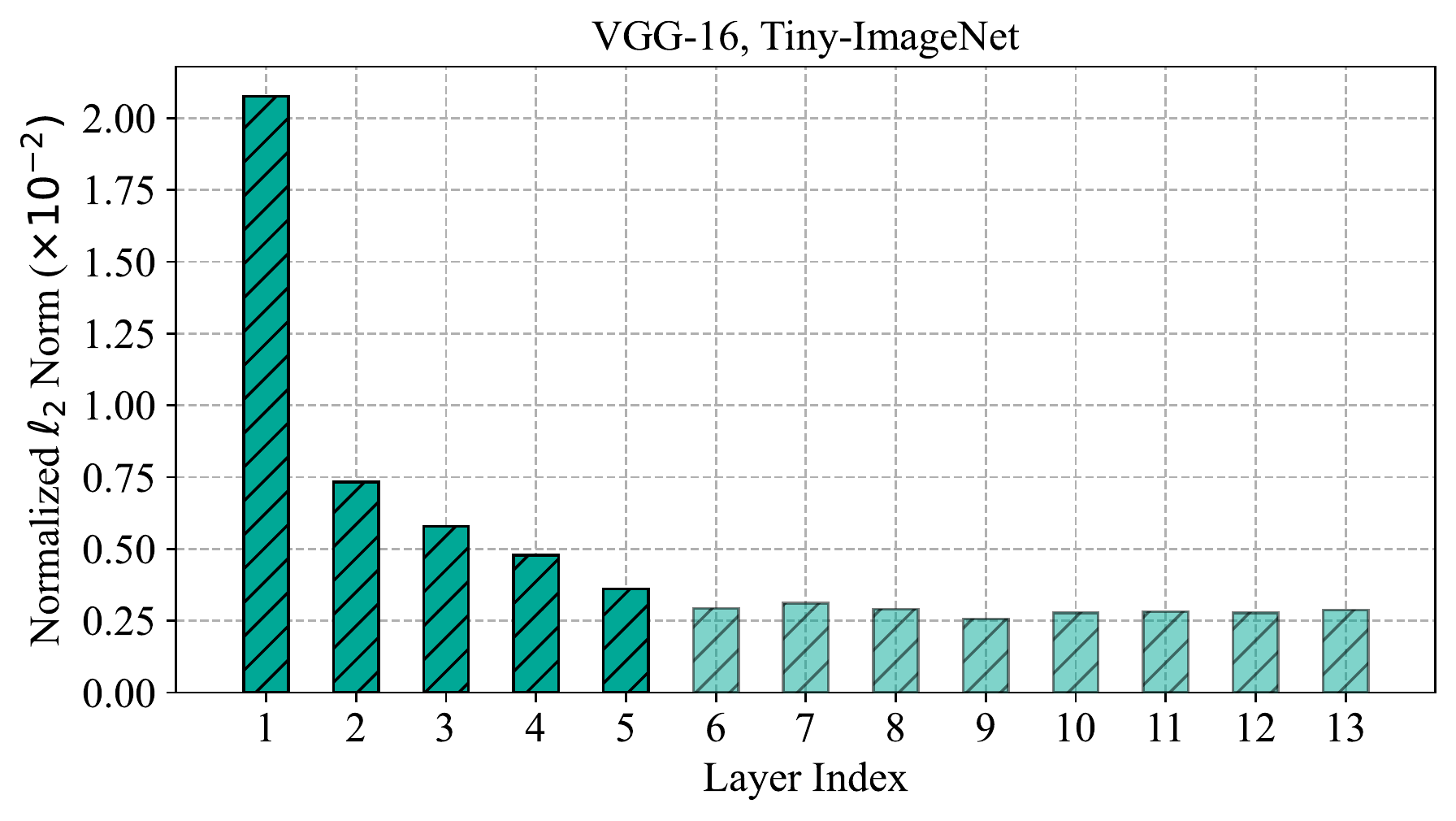}
     \vspace{-1.5em}
    \caption{On TinyImageNet and VGG-16, we visualize $\ell_2$ norms of RWSs for each convolutional layer, indicating the non-robust models and robust models mainly differ in the shallow layers.}
     \vspace{-0.5em}
    \label{fig:sec2:shallow_norm}
\end{figure}

\begin{figure*}[htb]
    \centering
    \includegraphics[width=0.9\linewidth]{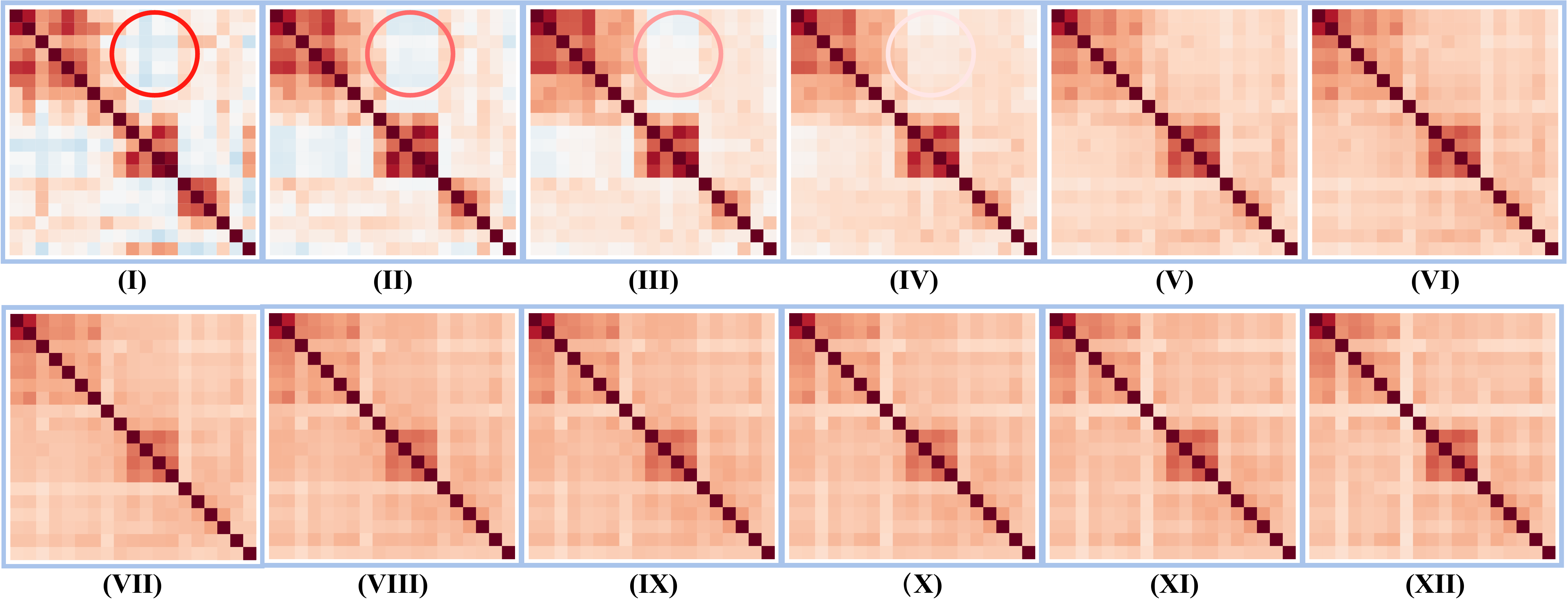}
        \vspace{-0.5em}
    \caption{Based on TinyImageNet and VGG-16, we visualize cosine similarities between different types of corruptions, at different layers. Lighter colors indicate smaller cosine similarities. RWSs of different corruption types are significantly more diverse in shallow layers.} 
    \label{fig:sec2:shallow_diversity}
    \vspace{-0.5em}
\end{figure*}

\section{Robust Weight Signatures: Concept Proofs}
\subsection{Definition and Notations}\label{sec2:definition}
 \vspace{-0.2em}
We compare non-robust and robust models in the weight space, to investigate how robustness is encoded. %
To begin with, model providers train a standard model $\theta_\text{std} \in \mathbb{R}^d$ on a clean dataset, and multiple robust counterparts $\theta_r^{c} \in \mathbb{R}^d$, each with corruption type $c$, from the same initialization $\theta_\text{init} \in \mathbb{R}^d$ used by $\theta_\text{std}$. 
We denote $\theta_\text{std}-\theta_\text{init}$ as the \textit{base direction} $v_\text{base}$, which contains knowledge of fitting standard dataset. For each corruption type $c$, we similarly compute the \textit{robustifying direction} $v_c = \theta_r^{c}-\theta_\text{init}$, which is assumed to contain the knowledge of resilience against corruption $c$. We then disentangle the robustness-part knowledge with the standard dataset-fitting knowledge, by subtracting $v_c$'s \textbf{projection} on $v_\text{base}$, from $v_c$ itself. We refer to the obtained residual vector as a robust weight signature (\textbf{RWS}):
\begin{equation}
    \label{equation:formulation}
    \mathbf{RWS}_c = v_c - P_{{v_{base}}}(v_c),
\end{equation}
where $P_{{v_{base}}}(v_c)$ denotes the projection operator from $v_c$ onto the column space of $v_{base}$, implemented by matrix pseudoinverse. The process of extracting RWS is also illustrated in the upper Figure~\ref{fig:teasor}.

Note that the projection-residual idea implies the (somewhat gross) assumption that the ``standard fitting knowledge" and ``robustness knowledge" are encoded \textbf{nearly orthogonally} in the robust model weights. The use of projection $P_{{v_{base}}}$ also \textbf{goes beyond} the vanilla weight arithmetic regime in \cite{ilharco2022editing} who simply subtract one weight from the other: we also tried the same and find it unable to extract effective RWSs (especially poor in composition). We conjecture that is because the weight gap between \{pre-trained, fine-tuned\} models \cite{ilharco2022editing} is either smaller or more linearly connected, compared to the weight gap between \{standard, robust\} models in our case. We leave the verification of those two open thoughts as future work. 

Besides the above, we point out two other substantial differences between our RWSs and ``task vectors" in \cite{ilharco2022editing}. Firstly, RWSs exhibit a compact and compressible structure (Sec. \ref{sec:shallow_layers}) which was not observed in task arithmetic \cite{ilharco2022editing}. Secondly, we observe RWSs to be consistent and transferrable across datasets (Sec \ref{sec2:consistent_across_dataset}), echoing our conjecture that robustness is perhaps encoded relatively independently of standard dataset content: the finding has no counterpart in \cite{ilharco2022editing} either.


\vspace{-0.5em}
\paragraph{Experimental Details.} We use three datasets, CIFAR-10, CIFAR-100 \cite{krizhevsky2009learning} and Tiny-ImageNet \cite{tiny-imagenet}, with two model architectures, VGG-16 \cite{simonyan2014very} and ResNet-50 \cite{he2016deep}. By default, we obtain a robust model, by training with the corresponding type of data augmentations applied to the clean dataset. All VGG-16 models use a learning rate of 0.01, while all ResNet-50 models use 0.001. We follow the corruption types in \cite{hendrycks2019robustness} and the corruption severity levels are set to be 5 (strongest) for all experiments by default. Intentionally, neither adversarial training nor more compositional augmentation was involved, because we want to ``purify" each RWS to cater to one corruption type, facilitating our later experiments to demonstrate their controllability and composition.

For the choice of common initialization $\theta_\text{init}$, we found that the same random initialization did not suffice to manifest the RWS phenomenon. That is understandable: compared to fine-tuning the same pre-trained model \cite{ilharco2022editing}, two models trained (standard or robustly) from scratch could be far away in their weight space, even using the same initialization and dataset, due to many randomness factors in the much longer training process. To ``anchor" the standard and robust model weights to be meaningfully close for RWS extraction, we explored two strategies: (1) use an ImageNet pre-trained model\footnote{https://pytorch.org/vision/stable/models.html} as $\theta_\text{init}$, and train both standard and robust models from there; (2) first train a standard model from scratch, and use it as $\theta_\text{init}$ to train all other robust models from. Both are found to expose RWSs much better and more stably, and we report results from the first option by default due to its superior cross-dataset transferrability. 


 \vspace{-0.2em}
\subsection{RWSs are Concentrated in Shallow Layers}\label{sec:shallow_layers}
 \vspace{-0.2em}
Intuitively, many image corruption artifacts interfere with the low-level features, inviting the natural guess: \textit{whether the corruption fragility of standard models, and correspondingly the robustness to them encoded by robust models, are mainly encoded in shallow layers}. Prior works have presented relevant findings. For example, \cite{huang2021exploring} observed that more parameters can improve robustness only when added to the shallow layers. We experimentally validate the hypothesis to be explicitly true.


Figures~\ref{fig:sec2:shallow_norm} and \ref{fig:sec2:shallow_diversity} visualize the norms (normalized to the same layer's standard weight norm, averaged across all corruption types) and diversities (cosine similarity across different corruption types) of RWSs from each layer. Overall, RWSs at shallower layers are (i) significantly larger in norm. For example, the first five layers occupy more than 65\% of total norm energy for RWSs extracted from VGG-16 on Tiny-Imagenet; (ii) significantly more diverse and discriminative between corruption types. Both observations imply that RWSs are more ``informative" in shallow layers. 

In all experiments hereinafter, we use RWSs in the \textbf{shallowest five layers} by default. This also leads us to aggressively compress RWSs in Sec~\ref{sec3:lightweight} for lightweight patching. 





 \vspace{-0.2em}
\subsection{RWSs Recover Corruption Relationships}\label{sec2:similar}
 \vspace{-0.2em}
We now take a deeper dive from Figure \ref{fig:sec2:shallow_diversity}, noting that different natural corruption types are not irrelevant. Instead, corruptions are roughly categorized into four groups: noise, blur, weather and digital \cite{hendrycks2019robustness}. \citep{yin2019fourier} also found that different corruptions related to different frequency domains. Corruptions within the same group category or frequency range are more similar, and the robustness against one corruption tends to help defend its similar ones too. On the contrary, different categories of corruptions may even offset each other's robustness.

\begin{figure}[htb]
    \centering
    \includegraphics[width=\linewidth]{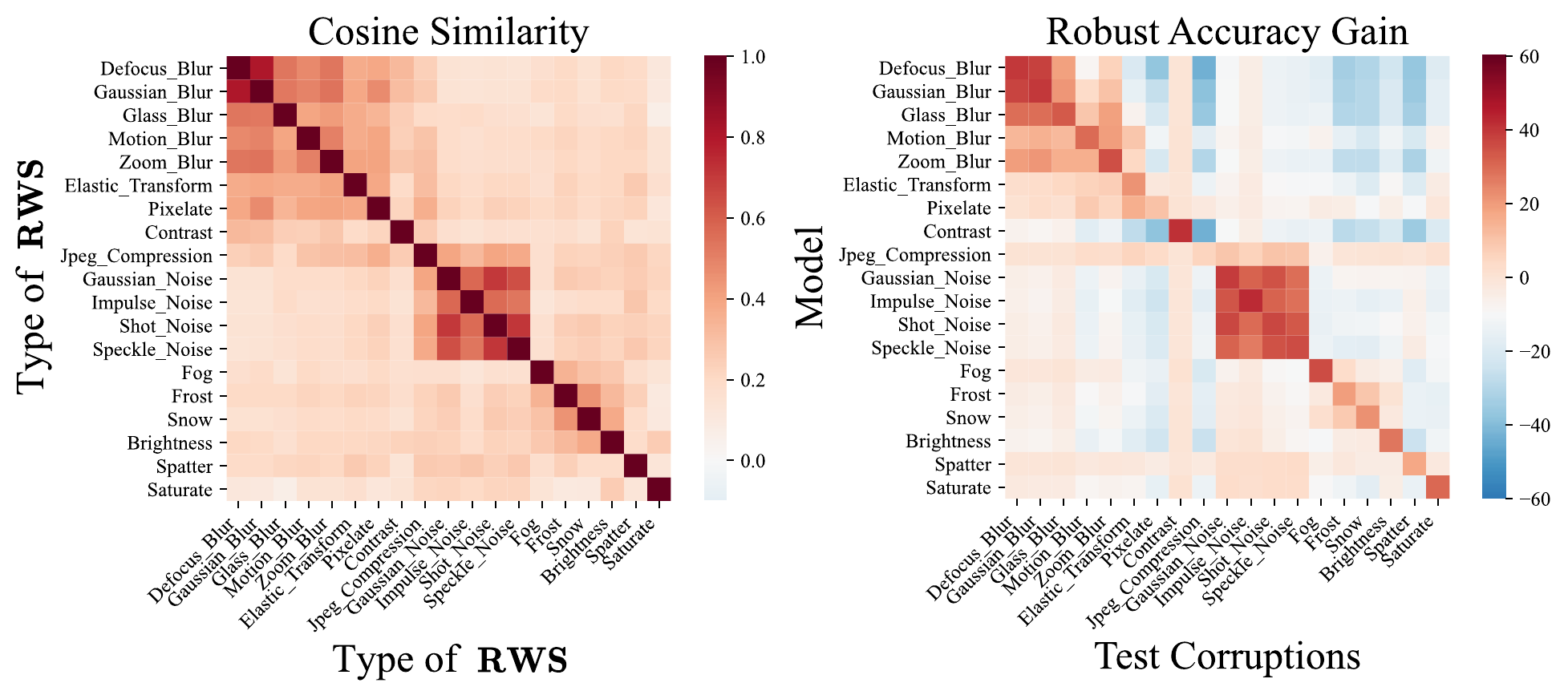}
    \vspace{-1.5em}
    \caption{Left: cosine similarities between RWSs of different corruption types. Right: the robust accuracy gains to all other corruptions, between a robust model trained solely on one corruption type and then tested on other types, and a standard model directly applied. For example, each element in the row `defocus blur' denotes the robust model trained with defocus blur and tested on other corruption types (column) - how much accuracy improvement or loss it will exhibit compared to the standard model. We use TinyImageNet with VGG-16.
}\vspace{-0.2em}
    \label{fig:sec2:relationship}
\end{figure}

Interestingly, RWSs successfully recover the relations of different corruption types. Figure~\ref{fig:sec2:relationship} (left) visualizes the cosine similarities between RWS of different corruption types, which reflect the grouping identified in \cite{hendrycks2019robustness}. We also visualize the robust accuracy gains to all other corruptions, when a robust model is trained solely on one corruption type and then directly tested on other types: the results in Figure~\ref{fig:sec2:relationship} (right) echo the former.

\begin{figure}[htb]
    \centering
    \includegraphics[width=\linewidth]{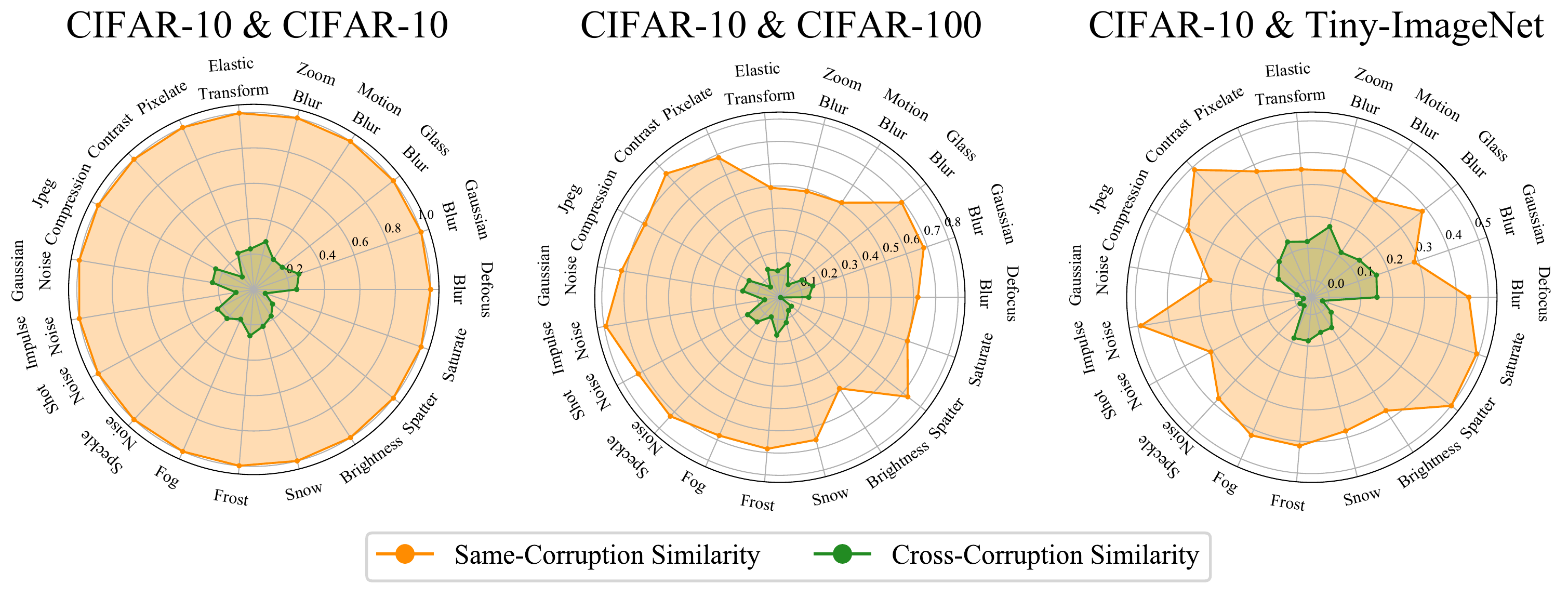}
    \vspace{-1em}
    \caption{Same-corruption (orange circles) and cross-corruption cosine similarities (green circles) between RWSs extracted from: (left) CIFAR-10 \& CIFAR-10; (middle) CIFAR-10 \& CIFAR-100; (right) CIFAR-10 \& TinyImageNet. The same-corruption similarity is computed between RWSs found on two datasets but of the same corruption type.  The cross-corruption similarity is computed as the average of cosine similarities between the current corruption type's RWS, and every other type's RWS.
    }\vspace{-1em}
    \label{fig:sec2:across_dataset}
\end{figure}

 \vspace{-0.2em}
\subsection{RWSs are Relatively Consistent across Datasets}\label{sec2:consistent_across_dataset}
 \vspace{-0.2em}
Now one more step further: we compare RWSs generated from different datasets. We are hopeful because of the (gross) assumption made back in Sec. \ref{sec2:definition}: the standard fitting and the robustness are relatively independent in weights. Figure~\ref{fig:sec2:across_dataset} partially confirms our hypothesis that RWS found from different datasets are relatively consistent. 

We first notice the same-corruption RWS cosine similarities across datasets to be consistently high. For example, between CIFAR-10 and CIFAR-100 (Figure~\ref{fig:sec2:across_dataset} middle subfigure), all same-corruption similarities are larger than 0.5, and some reach 0.8. Even comparing CIFAR-10 and Tiny-ImageNet (right) whose dataset statistics vary a lot, the same-corruption similarities are still all above 0.3 and sometimes reach 0.5. That implies the potential existence of ``universal model robustifying directions" which is even agnostic to standard model weights. 

On the other hand, the cross-corruption similarities remain consistent in the value range (most between 0.1 and 0.2), and more importantly, seem to preserve the relative similarity ``ranking" to some extent. For example, `Impulse noise' and `Saturate' have constantly the lowest cross-corruption similarities with others, while `Zoom blur' `Forst' `Defocus Blur'`Gaussian Blur' and `JPEG compression" are some of the consistent top rankers. We should note that this ranking consistency is imperfect: for example, `Gaussin Noise' is a high-ranker in the left and middle subfigures, but low on the right; while `Contrast' makes a vice versa case.


\vspace{-0.5em}
\section{An In-Situ Robustness Patching Framework}
\vspace{-0.2em}
Back to the main problem: \textit{how to achieve in-situ robustness?}
The aforementioned characteristics indicate the RWS involves discriminative and generalizable robust features, lending itself a promising option for direct weight patching on non-robust standard models. Given a standard model (trained on clean data) and its multiple robust counterparts (each trained on a corrupted version of the same dataset), we can store a single standard model and multiple RWSs. When needed, the robustness patching could be done immediately, by adding an RWS on standard model weight to create a \textit{patched} model $\theta_\text{patch}^c$ with extra robustness on corruption $c$:
\begin{equation}
    \theta_\text{patch}^c = \theta_\text{std} + \alpha * \textbf{RWS}_c
    \label{equation:patch}
\end{equation}
The appended RWS can be taken off any time to switch back to the intact standard model: hence there is no compromise of standard accuracy. $\alpha$ is a coefficient to adjust the ``strength" of the added robustness (Sec. \ref{sec3:adjust}), and the above equation could be extended to the weighted composition of multiple $\theta_\text{patch}^c$s with different corruptions $c$ (Sec. \ref{sec3:compose}).

\vspace{-0.2em}
\paragraph{More Related Work on ``Patching"} We shall credit existing literature that has studied model patching or similar notions. In general, many efforts have been invested to efficiently for altering a model’s behavior with post-training interventions, but without re-training. This stream of work may bear various names, such as patching \cite{goel2020model,ilharco2022patching,murty2022fixing}, editing \cite{mitchell2021fast,mitchell2022memory,santurkar2021editing}, aligning \cite{askell2021general,kasirzadeh2022conversation,ouyang2022training}, debugging \cite{geva2022lm,ribeiro2022adaptive,ilyas2022datamodels}, steering \cite{subramani2022extracting}, or reprogramming \cite{elsayed2018adversarial,tsai2020transfer,hambardzumyan2021warp,zhang2022fairness}. Those can operate on input, output, or weight levels, and many will take extra training or optimization steps. Several of them explored weight interpolation between a pre-trained model and its fine-tuned version, to either improve the fine-tuned model's distributional shift robustness \cite{wortsman2022robust}, or learn new specific tasks better without affecting other learned tasks \cite{ilharco2022patching}. 

The most relevant work to us is the task vector arithmetic \cite{ilharco2022editing}, which uniquely adds, scales, deletes or composes model capabilities, by applying vectors in the weight space of pre-trained models. They method is modular and efficient by re-using fine-tuned models, and does not modify the
standard fine-tuning procedure. However, \cite{ilharco2022editing} as well as \cite{wortsman2022robust,ilharco2022patching} focus on the fine-tuning setting and rely on large pre-trained models, while ours dig into a brand-new context. In Sec. \ref{sec2:definition}, we have also explained a few more differences between RWSs and task vectors in \cite{ilharco2022editing}, in both methodology and key findings. 

Next, we present a series of experiments to demonstrate the key advantages of our patching, namely, \textbf{lightweight}, \textbf{in-situ adjustable}, \textbf{composable}, and \textbf{transferrable}.



\vspace{-0.2em}
\subsection{Lightweight}\label{sec3:lightweight}
\vspace{-0.2em}
The first sanity check question is: why not store multiple robust models directly, but their weight differences? The answer: those differences are much more compressible and incur much less storage overhead. 
The storage efficiency of RWS is achieved by two aspects: ($1$) as analyzed in Sec~\ref{sec:shallow_layers}, we only use RWSs shallow layers, which usually contain much fewer parameters than latter layers; ($2$) we verify that RWS can further be compressed by quantization. 

We follow the same setting in Sec~\ref{sec2:definition} to construct RWSs and then follow Equation~\ref{equation:patch} to robustify the model. We set $\alpha$ as 1 by default. Our results are presented in Table~\ref{table:lightweight}. We provide several RWS options for patching standard models, including: 
($1$) $\textbf{RWS}_\text{full}$: RWSs from all layers are used. ($2$) $\textbf{RWS}_\text{shallow}$: RWSs are only kept from the shallowest five layers (default). ($3$) $\textbf{RWS}_\text{shallow,16bit}$: $\textbf{RWS}_\text{shallow}$ further quantized to 16 bit. ($4$) $\textbf{RWS}_\text{shallow,8bit}$: $\textbf{RWS}_\text{shallow}$ further quantized to 8 bit. We also include three baselines: `Standard' is the model trained on clean data only; `Data Augmentation' is the robust model trained with all 19 corruption types seen as training data augmentations; and `All Models' denotes the ensemble option, i.e., storing the standard model as well as 19 robust models (each dedicatedly trained with one corruption type) together. Note that `All Models' baseline assumes always using the right dedicated model in each situation (clean, or one of the 19 corrupted). Hence it effectively makes the performance ``upper bound" for all methods, though at the heaviest storage overhead. Meanwhile, `Data Augmentation' substantially boosts the corruption robustness without any storage overhead, but sacrifices the clean data performance meanwhile.

All RWS variations show significant effectiveness in robustifying standard models while retaining/recovering the standard accuracy when taking off RWSs. $\textbf{RWS}_\text{shallow,16bit}$ achieve a decent trade-off between storage cost and robustness; with only $20\% \sim 40\%$ storage cost increment than `Standard' or `Data Augmention', the method is able to (1) improve $30\% \sim 88\%$ averaged robustness gain across four cases, compared to the standard baseline; and (3) consistently outperform the ``Data Augmentation' baseline in achievable TA-RA trade-offs, 
$\textbf{RWS}_\text{shallow,8bit}$ further boosts the storage efficiency with small RA losses from 16-bit (especially, negligible on ResNet-50 + Tiny-ImageNet). More detailed comparisons and baseline results are in \underline{Appendix}. 
\begin{figure}[htb]
    \centering
    \includegraphics[width=\linewidth]{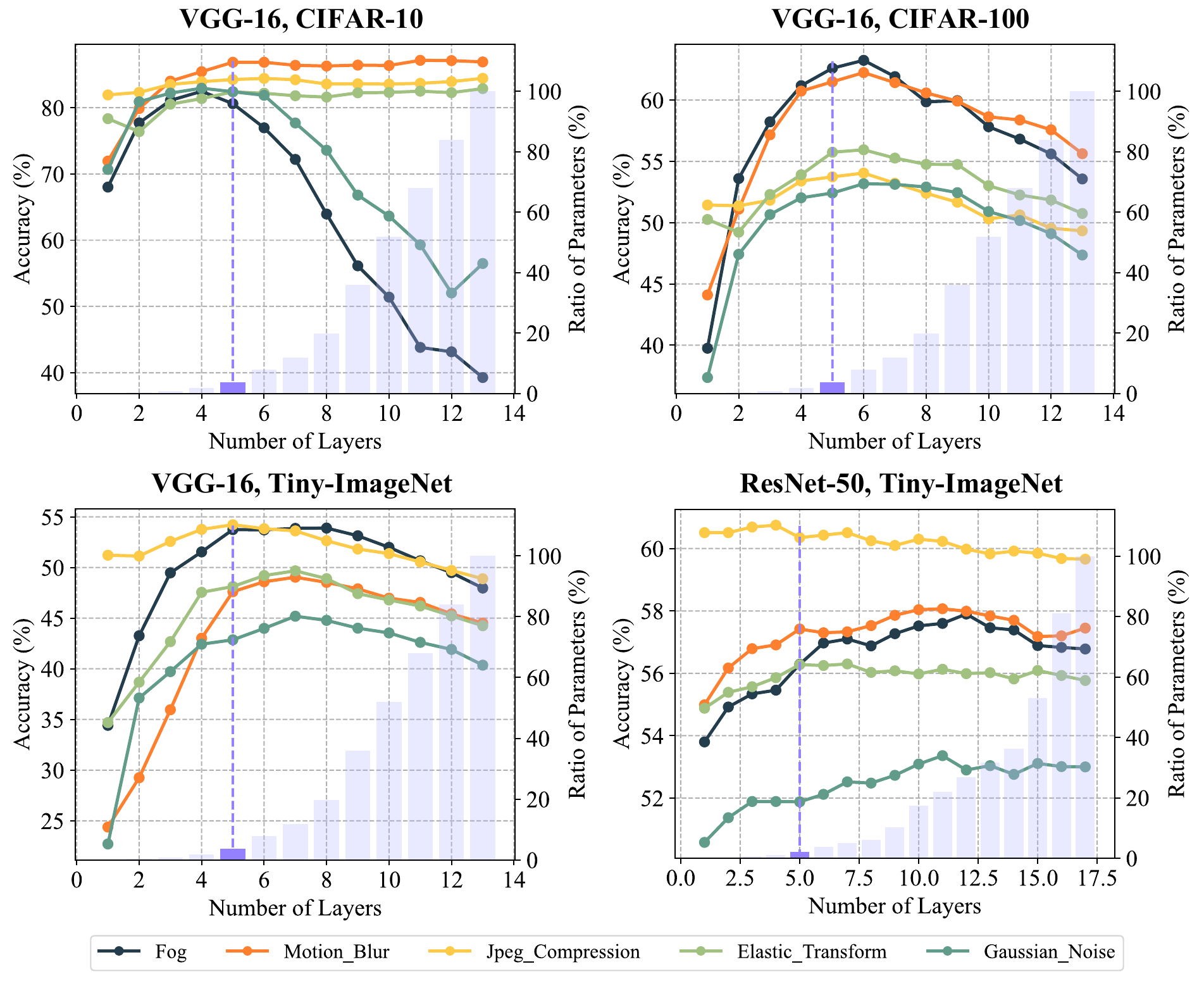}
    \vspace{-1.2em}
    \caption{Robustness trends (we select a few representative corruptions types) when altering the number of shallowest layers used for RWS construction. We also plot the used parameter ratios. 
    }
        \vspace{-1em}
    \label{fig:num_of_layer}
\end{figure}

\definecolor{LightCyan}{rgb}{0.88,1,1}
\newcolumntype{a}{>{\columncolor{LightCyan}}c}

\begin{table*}[htb]
\caption{Comparison of RWS-based methods and other options. We consider 19 corruption types in \cite{hendrycks2019robustness}. ``RA'' refers to averaged accuracy on all kinds of corrupted data, while ``TA'' refers to the test accuracy in the standard setting. $N_\mathrm{param}$ denotes the total model storage size (in MBs). Note that in the RWS-based pipeline, we report ``TA" for the model when the RWSs are taken off (hence fully recovering the standard model); and report ``RA" when the corresponding RWS is patched per corruption. This is an ideal case of using RWS patching - and its rationale and limitations will be both discussed in Sec. \ref{sec4:limit}. Similarly, as a fair comparison, the `All Models' baseline assumes we always use the right dedicated standard/robust model in each clean/corrupted situation, too. \label{table:lightweight}}
\resizebox{1\linewidth}{!}{
\begin{tabular}{l|cca|cca|cca|cca}
\toprule
\multirow{3}{*}{\textbf{Methods}} & \multicolumn{3}{c|}{CIFAR-10} & \multicolumn{3}{c|}{CIFAR-100} & \multicolumn{6}{c}{Tiny-ImageNet} \\\cmidrule{2-13}
& \multicolumn{3}{c|}{VGG-16} & \multicolumn{3}{c|}{VGG-16} & \multicolumn{3}{c|}{VGG-16} & \multicolumn{3}{c}{ResNet-50} \\ \cmidrule{2-13}
& $N_\mathrm{param}$ (MB) & TA (\%) & RA (\%) & $N_\mathrm{param}$ (MB) & TA (\%) & RA (\%) &$N_\mathrm{param}$ (MB) & TA (\%) & RA (\%) &$N_\mathrm{param}$ (MB) & TA (\%) & RA (\%)  \\ \midrule

Standard & $58.8$ & $92.59$ & $65.44$ & $58.8$ & $71.44$ & $36.76$ & $59.2$ & $61.28$ & $23.58$ & $95.7$ & $65.72$ & $29.65$  \\ \midrule

Data Augmentation & $58.8$ & $89.58$ & $84.34$ & $58.8$ & $67.34$ & $56.95$ & $59.2$ & $52.11$ & $43.64$ & $95.7$ & $59.17$ & $47.96$  \\
All Models & $1177.6$ \textcolor{blue}{($20 \times$)} & $92.59$ & $88.97$ & $1177.6$ \textcolor{blue}{($20 \times$)} & $71.44$ & $64.72$ & $1184.0$ \textcolor{blue}{($20 \times$)} & $61.28$ & $51.55$ & $1913.6$ \textcolor{blue}{($20 \times$)} & $65.72$ & $55.97$ \\\midrule

Standard+$\textbf{RWS}_\mathrm{Full}$ & $1177.6$ \textcolor{blue}{($20 \times$)} & $92.59$ & $75.35$ & $1177.6$ \textcolor{blue}{($20 \times$)} & $71.44$ & $52.58$ & $1184.0$ \textcolor{blue}{($20 \times$)} & $61.28$ & $43.63 $ & $1913.6$ \textcolor{blue}{($20 \times$)} & $65.72$ & $53.64$  \\
Standard+$\textbf{RWS}_\mathrm{Shallow}$ & $101.0$ \textcolor{blue}{($1.7 \times$)} & $92.59$ & $84.86$ & $101.0$ (\textcolor{blue}{$1.7 \times$)} & $71.44$ & $58.78$ & $101.4$ \textcolor{blue}{($1.7 \times$)} & $61.28$ & $44.66$ & $131.4$ \textcolor{blue}{($1.4 \times$)} & $65.72$ & $52.84$  \\
Standard+$\textbf{RWS}_\mathrm{Shallow, 16 bits}$ & $\textbf{79.9}$ (\textcolor{blue}{$\textbf{1.4} \times$)} & $\textbf{92.59}$ & $\textbf{84.76}$ & $\textbf{79.9}$ \textcolor{blue}{($\textbf{1.4} \times$)} & $\textbf{71.44}$ & $\textbf{58.62}$ & $\textbf{80.3}$ \textcolor{blue}{($\textbf{1.4} \times$)} & $\textbf{61.28}$ & $\textbf{44.25}$ & $\textbf{113.6}$ \textcolor{blue}{($\textbf{1.2} \times$)} & $\textbf{65.72}$ &$\textbf{52.81}$  \\
Standard+$\textbf{RWS}_\mathrm{Shallow, 8 bits}$ & $69.4$ \textcolor{blue}{($1.2 \times$)} & $92.59$ & $82.99$ & $69.4$ \textcolor{blue}{($1.2 \times$)} & $71.44$ & $53.52$ & $69.7$ \textcolor{blue}{($1.2 \times$)} & $61.28$ &$39.40$ & $104.7$ \textcolor{blue}{($1.1 \times$)} & $65.72$ &$52.79$ \\


 \bottomrule
\end{tabular}}
\end{table*}

We further alter the number of layers used for constructing RWS and plot the average robust accuracy of patched models (left bar of each subfigure), accompanied by the corresponding ratio of parameters participated (right bar). Figure~\ref{fig:num_of_layer} shows that: ($1$) the robustness of patched VGG models does not benefit from using more layers in extracting RWSs, and actually will be ``backfired" when more latter layers are included; ($2$) ResNet models also see saturation effects on the robustness of most corruption types, after more than 5 layers are used. Both imply the high-level features have little to do with robustness encoding, and justify our design choice of using only the shallowest few layers.

\begin{figure}[htb]
    \centering
    \includegraphics[width=\linewidth]{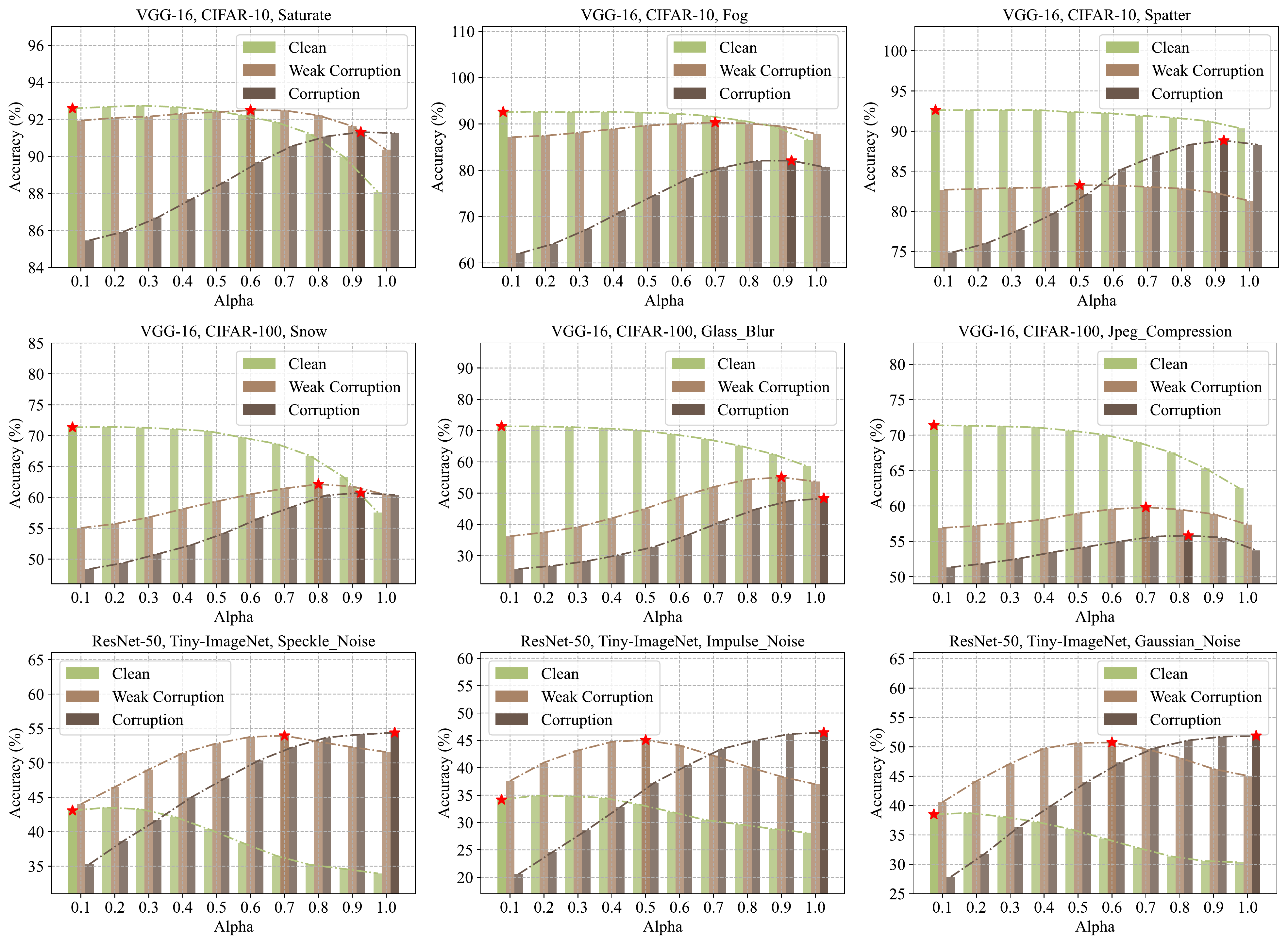}
    \vspace{-1em}
    \caption{Effect of $\alpha$ on the robustness of the patched model under different types and severity levels of corrupted data. \textcolor{ygreen}{Green} bars, \textcolor{ybrown}{brown} bars, \textcolor{ydarky}{dark brown} bars represent clean accuracy, robust accuracy under the corruption of severity level 3, and robust accuracy under the corruption of severity level 5, respectively.}\vspace{-1.5em}
    \label{fig:adjustable}
\end{figure}

\begin{figure}[htb]
    \centering
    \vspace{-0.5em}
    \includegraphics[width=0.9\linewidth]{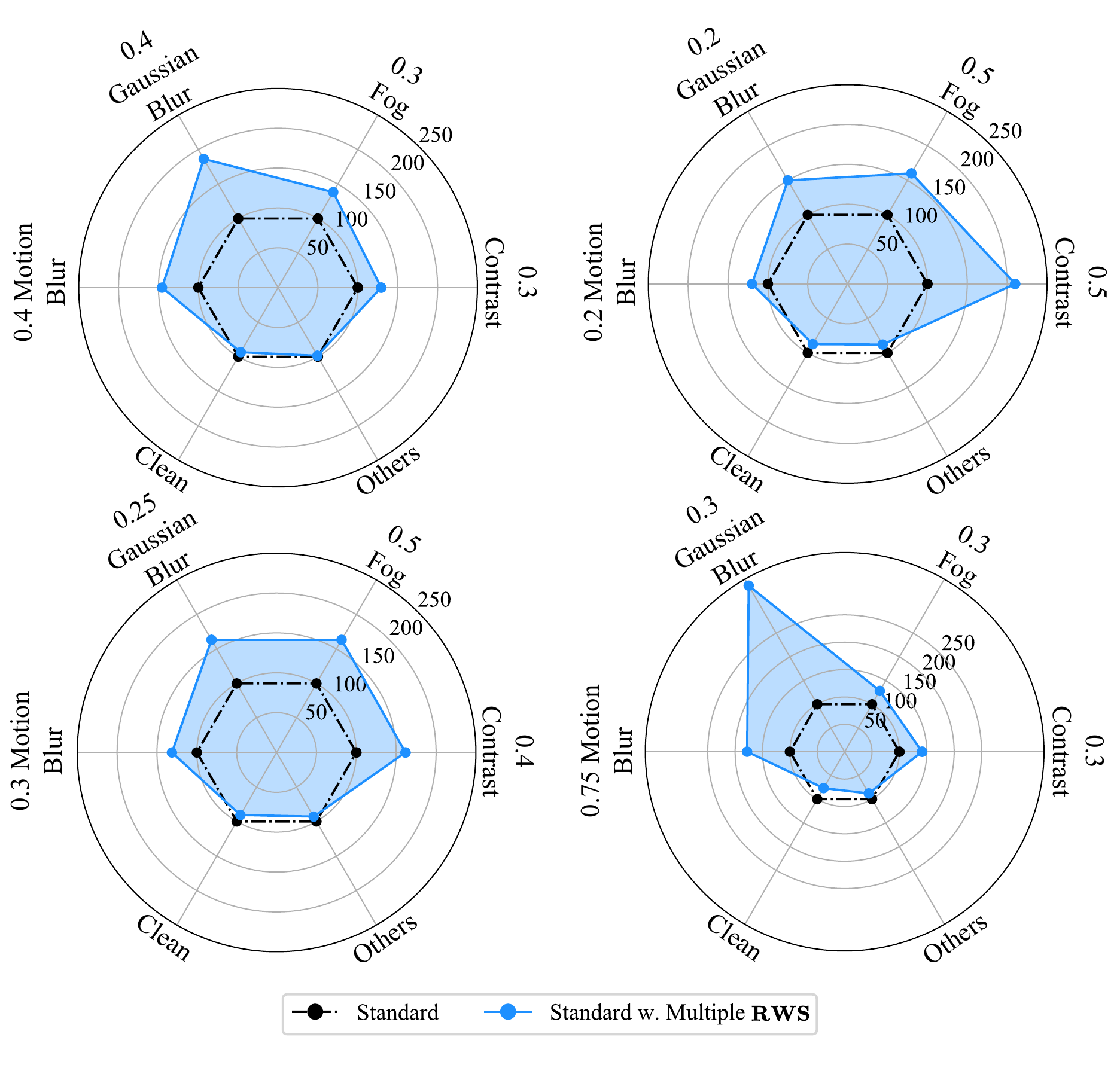}
    \vspace{-1.8em}
    \caption{Visualization of robustness improvements when adding multiple RWSs together. On VGG-16 and Tiny-ImageNet, we add 4 different RWSs of ``motion blur'', ``gaussian blur'', ``fog'' and ``contrast'' together. By changing their coefficients, we can obtain robust models with different specialties, with minimal loss of clean accuracy and robustness on other corruption types.}\vspace{-1.2em}
    \label{fig:composable}
\end{figure}

\subsection{In-Situ Adjustable} \label{sec3:adjust}
\vspace{-0.2em}
RWSs can be not only applied to patch a single robust model per corruption: they can even adapt to any corruption levels (e.g. different visibility in the fog weather) by linearly re-scaling, easily achieving the smooth trade-off between standard and robustness performances in \underline{one same model} (note this is different from adding/taking off an RWS, which is essentially switching between two models). This can be achieved by adjusting the coefficient $\alpha \in [0,1]$ in Equation~\ref{equation:patch}: essentially, that is interpolating the (shallow layers') weights between the standard and robust models.

To validate, we test the patched model on corrupted data with different severity levels (as defined in \cite{hendrycks2019benchmarking}). As in Figure \ref{fig:adjustable}, for instance on CIFAR-10 and VGG-16, patched models always achieve the best performance at the severity level 5 (strongest corruptions) when $\alpha=0.9$ or $1$. When the severity level is set to 3, the patched model performs the best when $\alpha=0.6$. Meanwhile, the standard accuracy gracefully decays as $\alpha$ increases. 


\begin{figure*}[htb]
    \centering
    \includegraphics[width=0.9\linewidth]{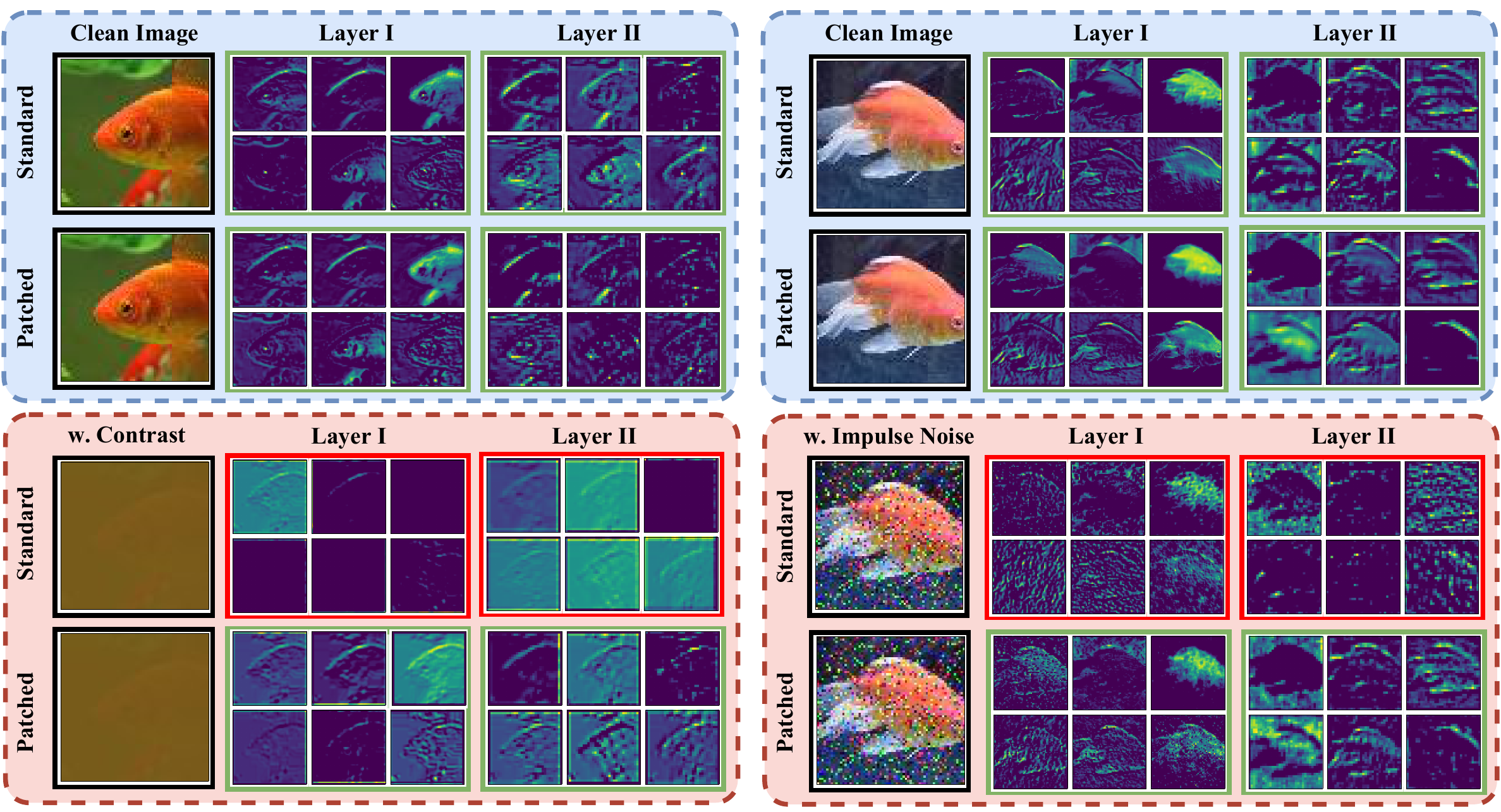}
    \caption{The comparison of the patched model and the standard model's feature maps given the same input sample shows our RWS patching method meaningfully equips the model with resistance to different types of corruptions.}
     \vspace{-0.8em}
    \label{fig:visualization}
\end{figure*}

\vspace{-0.2em}
\subsection{Composable}\label{sec3:compose}
\vspace{-0.2em}
Usually, images usually do not just suffer from a single type of corruption. To resist compound natural corruptions, 
RWSs can also be linearly composed to form a model of multi-corruption robustness, by extending Equation~\ref{equation:patch} to adding multiple RWSs, each with their own $\alpha$s. The previous in-situ adjustment could also be seen as a special case. We can even control the linear combination coefficient to obtain models with different ``robustness specialties". Figure~\ref{fig:composable} shows that by composing RWSs with different coefficients, one can construct a wide range of models with different strengths at simultaneously tackling diverse corruptions. That leads us to an ``infinite pool" of possible models, by just re-composing a small pool of RWSs and no re-training.


The composable property of RWSs reminds the weight interpolation between two different models \cite{izmailov2018averaging,zhao2020bridging,ilharco2022patching,wortsman2022model,wortsman2022robust,choshen2022fusing}, yet composing natural corruption robustness seems a new theme. Note that though, the construction of RWSs needs to first remove the robust weight's projection onto the standard weight column space, hence composing RWSs does not naively equal interpolating their source robust model weights.

\vspace{-0.2em}
\subsection{Transferable}
\vspace{-0.2em}
Lastly, the cross-dataset consistency of RWSs as analyzed in Sec~\ref{sec2:consistent_across_dataset} motivates us to study if an RWS found from one dataset can be reused for the same architecture trained on a different dataset, to transfer robustness to the latter ``for free". 
Table~\ref{table:transferability} confirms this possibility. Despite the training data domain shift of the standard model, RWSs stay effective for patching robustness in a ``zero-shot" manner. Unsurprisingly also, smaller gaps will render the RWS transfer more effective. For example, the robustness gain of CIFAR-100 by patching CIFAR-10 RWSs is clearly larger than the Tiny-ImageNet gain by patching the same.

In addition, the strong transferability implies the tantalizing possibility of gaining robustness efficiently by ``transferring" RWSs from small to large datasets. Specifically, direct robust training with data augmentations on large datasets such as ImageNet can be resource-demanding. Instead, one can first extract RWSs by robust-training over smaller datasets (e.g. CIFAR-10 or TinyImageNet), and subsequently, transfer them to ``patching" the same model architecture standard-trained on the target large dataset. The results, as presented in Table~\ref{table:efficient_surrogate}, demonstrate that the ``out of the box" application of RWS can lead to significant gains in ImageNet robustness, especially when RWS is obtained from TinyImageNet (whose distribution is the most similar to ImageNet). 

\vspace{-0.2em}
\subsection{Feature Map Visualization}
\vspace{-0.2em}
Besides, we compare feature maps of standard and patched models in Figure~\ref{fig:visualization}, to understand what information is actually patched. Using TinyImageNet and VGG-16, we visualize feature maps after the second and third convolutional layers (denoted as ``Layer \uppercase\expandafter{\romannumeral1}'' and ``Layer \uppercase\expandafter{\romannumeral2}'', respectively). The visualizations show that RWSs bring in meaningful feature adjustments to be resilient to corruption types. For example, to tackle reduced contrast, the patched model becomes more sensitive to edges, while the model patched for impulse noise picks up less high-frequency outlier features. 

\setlength{\columnsep}{4pt}%
\setlength{\intextsep}{4pt}%
\begin{table}
\centering
\vspace{-1em}
\caption{\small Robustness gains when applying RWSs extracted from small datasets (CIFAR-10, CIFAR-100, Tiny-ImageNet) to full ImageNet models. Robust accuracy is evaluated on ImageNet and averaged on all kinds of corrupted data. All models use the VGG-16 architecture. $\mathrm{Data Augmentation}^{1}$ use the same FLOPS as the overhead of extracting RWS based on CIFAR-10 models, while $\mathrm{Data Augmentation}^{2}$ use the same FLOPS as the RWS extraction based on TinyImageNet models.}\label{table:efficient_surrogate}
\resizebox{0.9\linewidth}{!}{
\begin{tabular}{l|c}
\toprule
Method & Robust Accuracy ($\%$)  \\
\midrule
Standard & $11.14$ \\
\midrule
$\mathrm{Data Augmentation}^{1}$ & $11.01$ \textcolor{blue}{($\downarrow 0.13$)}\\
$\mathrm{Data Augmentation}^{2}$ & $14.52$ \textcolor{blue}{($\uparrow 3.38$)} \\
\midrule
Standard + $\textbf{RWS}_{\mathrm{CIFAR}-10}$ & $13.47$ \textcolor{blue}{($\uparrow 2.33$)} \\
Standard + $\textbf{RWS}_{\mathrm{CIFAR}-100}$ & $14.55$ \textcolor{blue}{($\uparrow 3.41$)} \\
Standard + $\textbf{RWS}_{\mathrm{Tiny-Imagenet}}$ & $\textbf{17.53}$ \textcolor{blue}{($\uparrow \textbf{6.39}$)} \\

\bottomrule
\end{tabular}}
\vspace{-0.5em}
\end{table}

\definecolor{LightCyan}{rgb}{0.88,1,1}

\begin{table}[htb]
\centering
\vspace{-0.5em}
\caption{Transferring RWSs across datasets. For three non-robust models trained on CIFAR-10, CIFAR-100 and Tiny-ImageNet (by columns), we patch RWSs generated from CIFAR-10, CIFAR-100, and Tiny-ImageNet (by rows), respectively, for injecting zero-shot robustness. The robust accuracies are averaged across 19 corruptions/19 RWSs. We use the VGG-16 model here. \label{table:transferability}}
\resizebox{1.0\linewidth}{!}{
\begin{tabular}{l|ccc}
\toprule
\multirow{2}{*}{Methods} & \multicolumn{3}{c}{Robust Accuracy ($\%$)}\\
\cmidrule{2-4}
& CIFAR-10 & CIFAR-100 & Tiny-ImageNet \\ \midrule
Standard & $65.44$ & $36.76$ & $23.58$ \\ \midrule
Standard + $\textbf{RWS}_{\mathrm{CIFAR}-10}$ & $84.76$ \textcolor{blue}{($\uparrow 19.32$)} & $55.65$ \textcolor{blue}{($\uparrow 18.89$)} & $32.23$ \textcolor{blue}{($\uparrow 8.65$)} \\
Standard + $\textbf{RWS}_{\mathrm{CIFAR}-100}$ & $83.01$ \textcolor{blue}{($\uparrow 17.57$)} & $58.62$ \textcolor{blue}{($\uparrow 21.86$)} & $33.17$ \textcolor{blue}{($\uparrow 9.59$)} \\
Standard + $\textbf{RWS}_{\mathrm{Tiny}-\mathrm{ImageNet}}$ & $71.43$ \textcolor{blue}{($\uparrow 5.99$)} & $44.94$ \textcolor{blue}{($\uparrow 8.18$)} & $44.25$ \textcolor{blue}{($\uparrow 20.67$)} \\

\bottomrule
\end{tabular}}
\vspace{-0.5em}
\end{table}

\vspace{-0.5em}
\section{Conclusion and Limitations}\label{sec4:limit}
\vspace{-0.2em}
Our work is dedicated to investigating how natural corruption ``robustness'' is encoded in weights and how to disentangle/transfer them. We introduce \textit{``Robust Weight Signature''}(\textbf{RWS}), which nontrivially generalizes the prior wisdom in model weight interpolation and arithmetic, to analyzing standard/robust models, with both methodological innovations and new key findings. RWSs lead to a powerful in-situ model patching framework to easily achieve on-demand robustness towards a wide range of corruptions.

Current RWS patching faces one limitation that we must point out: in Table~\ref{table:lightweight}, the superior TA/RA trade-offs achieved by RWS methods are based on the perfect ``oracle" knowledge of what corruption is being handled, i.e., when to add or take off the ``correct" RWSs.  This assumption is in line with the `once-for-all" AT methods \cite{wang2020once,kundu2023float}, which requires a human oracle to control a test-time hyperparameter to implicitly state the desired RA-TA trade-offs in contexts. Practically, that can be implemented by referring to environment sensors or other domain classification change or detection methods. We also ensure our fair comparison with  ``All Models" baseline in Table \ref{table:lightweight} by using the same ideal oracle.  

One future work of immediate interest would be to examine RWS patching under a practical imperfect oracle (e.g., a trained corruption domain classifier that might predict incorrectly, hence applying inexact RWSs). We hypothesize the overall performance drop will be mild though, since an RWS trained for one corruption type can boost robustness against other ``similar" corruptions too (see Sec. \ref{sec2:similar}).




\bibliography{RWS}
\bibliographystyle{icml2023}

\newpage
\appendix
\onecolumn

\section{More Experimental Results}\label{sec:app:results}
\subsection{Detailed Results on All 19 Corruption Types}
To supplement Table \ref{table:lightweight}, we provide the detailed experimental results of all corruption types in Table~\ref{table:appendix:full_result}, which shows our consistent improvements in the RAs of all corruption types. We use $\textbf{RWS}_\text{shallow,16bit}$ for model patching: RWS are constructed by the shallowest five layers with 16-bit quantization.

\begin{table*}[htb]
\centering
\caption{Detailed experimental results showing robustness improvements on all 19 corruption types in \cite{hendrycks2019robustness}. \label{table:appendix:full_result}}
\resizebox{1\linewidth}{!}{
\begin{tabular}{c|ccl|ccl|ccl|ccl}
\toprule
\multirow{3}{*}{\textbf{Corruptions}} & \multicolumn{3}{c|}{CIFAR-10} & \multicolumn{3}{c|}{CIFAR-100} & \multicolumn{6}{c}{Tiny-ImageNet} \\\cmidrule{2-13}
& \multicolumn{3}{c|}{VGG-16} & \multicolumn{3}{c|}{VGG-16} & \multicolumn{3}{c|}{VGG-16} & \multicolumn{3}{c}{ResNet-50} \\ \cmidrule{2-13}
& w.o. RWS (\%) & w. RWS (\%) & Diff. & w.o. RWS (\%) & w. RWS (\%) & Diff. & w.o. RWS (\%) & w. RWS (\%) & Diff.& w.o. RWS (\%) & w. RWS (\%) & Diff.  \\ 
\midrule
Brightness & 88.41 & 90.87 & \textcolor{blue}{$\uparrow$ 2.46}   & 60.85 & 67.06 & \textcolor{blue}{$\uparrow$ 6.21}   & 30.33 & 49.41 & \textcolor{blue}{$\uparrow$ 19.08}   & 37.96 & 56.61 & \textcolor{blue}{$\uparrow$ 18.65} \\
Contrast & 41.29 & 78.77 & \textcolor{blue}{$\uparrow$ 37.48}   & 16.11 & 60.72 & \textcolor{blue}{$\uparrow$ 44.61}   & 1.88 & 19.30 & \textcolor{blue}{$\uparrow$ 17.42}   & 1.88 & 28.55 & \textcolor{blue}{$\uparrow$ 26.67} \\
Defocus Blur & 66.92 & 87.14 & \textcolor{blue}{$\uparrow$ 20.22}   & 37.55 & 61.51 & \textcolor{blue}{$\uparrow$ 23.96}   & 7.01 & 34.91 & \textcolor{blue}{$\uparrow$ 27.90}   & 26.09 & 51.03 & \textcolor{blue}{$\uparrow$ 24.94} \\
elastic Transform & 78.38 & 81.93 & \textcolor{blue}{$\uparrow$ 3.55}   & 49.74 & 56.01 & \textcolor{blue}{$\uparrow$ 6.27}   & 35.06 & 48.15 & \textcolor{blue}{$\uparrow$ 13.09}   & 41.60 & 56.12 & \textcolor{blue}{$\uparrow$ 14.52} \\
Fog & 61.40 & 80.57 & \textcolor{blue}{$\uparrow$ 19.17}   & 30.18 & 62.33 & \textcolor{blue}{$\uparrow$ 32.15}   & 27.59 & 53.51 & \textcolor{blue}{$\uparrow$ 25.92}   & 20.51 & 56.23 & \textcolor{blue}{$\uparrow$ 35.72} \\
Frost & 70.65 & 84.89 & \textcolor{blue}{$\uparrow$ 14.24}   & 40.73 & 56.80 & \textcolor{blue}{$\uparrow$ 16.07}   & 38.40 & 51.22 & \textcolor{blue}{$\uparrow$ 12.82}   & 42.23 & 56.47 & \textcolor{blue}{$\uparrow$ 14.24} \\
Gaussian Blur & 56.63 & 86.71 & \textcolor{blue}{$\uparrow$ 30.08}   & 30.80 & 60.91 & \textcolor{blue}{$\uparrow$ 30.11}   & 8.82 & 40.75 & \textcolor{blue}{$\uparrow$ 31.93}   & 29.28 & 54.07 & \textcolor{blue}{$\uparrow$ 24.79} \\
Gaussian Noise & 50.92 & 82.08 & \textcolor{blue}{$\uparrow$ 31.16}   & 24.79 & 52.83 & \textcolor{blue}{$\uparrow$ 28.04}   & 12.37 & 42.38 & \textcolor{blue}{$\uparrow$ 30.01}   & 15.29 & 51.86 & \textcolor{blue}{$\uparrow$ 36.57} \\
Glass Blur & 56.36 & 78.55 & \textcolor{blue}{$\uparrow$ 22.19}   & 25.54 & 48.48 & \textcolor{blue}{$\uparrow$ 22.94}   & 5.92 & 20.89 & \textcolor{blue}{$\uparrow$ 14.97}   & 15.39 & 39.11 & \textcolor{blue}{$\uparrow$ 23.72} \\
Impulse Noise & 39.82 & 77.24 & \textcolor{blue}{$\uparrow$ 37.42}   & 12.01 & 49.35 & \textcolor{blue}{$\uparrow$ 37.34}   & 6.36 & 37.77 & \textcolor{blue}{$\uparrow$ 31.41}   & 7.98 & 46.22 & \textcolor{blue}{$\uparrow$ 38.24} \\
Jpeg Compression & 80.72 & 84.26 & \textcolor{blue}{$\uparrow$ 3.54}   & 51.18 & 53.71 & \textcolor{blue}{$\uparrow$ 2.53}   & 51.23 & 54.16 & \textcolor{blue}{$\uparrow$ 2.93}   & 57.80 & 60.47 & \textcolor{blue}{$\uparrow$ 2.67} \\
Motion Blur & 68.98 & 86.88 & \textcolor{blue}{$\uparrow$ 17.90}   & 40.88 & 61.42 & \textcolor{blue}{$\uparrow$ 20.54}   & 23.63 & 47.35 & \textcolor{blue}{$\uparrow$ 23.72}   & 34.59 & 57.58 & \textcolor{blue}{$\uparrow$ 22.99} \\
Pixelate & 63.92 & 87.24 & \textcolor{blue}{$\uparrow$ 23.32}   & 36.09 & 62.05 & \textcolor{blue}{$\uparrow$ 25.96}   & 48.84 & 53.41 & \textcolor{blue}{$\uparrow$ 4.57}   & 53.48 & 60.51 & \textcolor{blue}{$\uparrow$ 7.03} \\
Saturate & 85.27 & 91.30 & \textcolor{blue}{$\uparrow$ 6.03}   & 53.55 & 66.47 & \textcolor{blue}{$\uparrow$ 12.92}   & 23.97 & 46.37 & \textcolor{blue}{$\uparrow$ 22.40}   & 29.70 & 52.73 & \textcolor{blue}{$\uparrow$ 23.03} \\
Shot Noise & 54.38 & 84.97 & \textcolor{blue}{$\uparrow$ 30.59}   & 27.57 & 54.81 & \textcolor{blue}{$\uparrow$ 27.24}   & 16.17 & 46.38 & \textcolor{blue}{$\uparrow$ 30.21}   & 18.74 & 53.98 & \textcolor{blue}{$\uparrow$ 35.24} \\
Snow & 78.42 & 87.49 & \textcolor{blue}{$\uparrow$ 9.07}   & 48.18 & 59.75 & \textcolor{blue}{$\uparrow$ 11.57}   & 34.92 & 50.94 & \textcolor{blue}{$\uparrow$ 16.02}   & 37.98 & 56.06 & \textcolor{blue}{$\uparrow$ 18.08} \\
Spatter & 74.36 & 88.18 & \textcolor{blue}{$\uparrow$ 13.82}   & 41.22 & 62.98 & \textcolor{blue}{$\uparrow$ 21.76}   & 40.99 & 55.31 & \textcolor{blue}{$\uparrow$ 14.32}   & 46.07 & 57.04 & \textcolor{blue}{$\uparrow$ 10.97} \\
Speckle Noise & 55.86 & 85.26 & \textcolor{blue}{$\uparrow$ 29.40}   & 27.81 & 54.82 & \textcolor{blue}{$\uparrow$ 27.01}   & 18.31 & 47.48 & \textcolor{blue}{$\uparrow$ 29.17}   & 20.28 & 54.39 & \textcolor{blue}{$\uparrow$ 34.11} \\
Zoom Blur & 70.63 & 86.02 & \textcolor{blue}{$\uparrow$ 15.39}   & 43.60 & 61.81 & \textcolor{blue}{$\uparrow$ 18.21}   & 16.20 & 41.13 & \textcolor{blue}{$\uparrow$ 24.93}   & 26.45 & 54.34 & \textcolor{blue}{$\uparrow$ 27.89} \\
\midrule
Average & 65.44 & 84.76 & \textcolor{blue}{$\uparrow$ 19.32} & 36.76 & 58.62 & \textcolor{blue}{$\uparrow$ 21.86} & 23.58 & 44.25 & \textcolor{blue}{$\uparrow$ 20.67} & 29.65 & 52.81 & \textcolor{blue}{$\uparrow$ 23.16}\\
\bottomrule 
\end{tabular}}
\end{table*}

\subsection{Compressibility: Full models versus RWSs under Quantization}

As is shown in Table~\ref{table:more_quantization}, full model weights are less compressible compared to RWSs, which suggests RWS-based methods to more easily achieve better storage efficiency. When applying the linear 16-bit quantization \cite{wu2020integer} to the full model weights, both the standard accuracy and natural corruption robustness have already degraded heavily. This is in stark contrast to RWSs which can retain most of their performance under the same quantization (16-bit) or even heavier (8-bit). 

Note that we focus our study to provide the proof of concept that ``RWSs are more easily amendable to quantization". We do not exclude the possibility that more sophisticated, robustness-aware quantization methods will sustain the robustness performance under heavy quantization, but further testing or developing such algorithms is out of this paper's scope.

\begin{table*}[htb]
\caption{The standard and robust accuracy changes when applying different levels of quantization, showing the superior compressiblity of RWS-based methods. \label{table:more_quantization}}
\resizebox{1\linewidth}{!}{
\begin{tabular}{l|cca|cca|cca|cca}
\toprule
\multirow{3}{*}{\textbf{Methods}} & \multicolumn{3}{c|}{CIFAR-10} & \multicolumn{3}{c|}{CIFAR-100} & \multicolumn{6}{c}{Tiny-ImageNet} \\\cmidrule{2-13}
& \multicolumn{3}{c|}{VGG-16} & \multicolumn{3}{c|}{VGG-16} & \multicolumn{3}{c|}{VGG-16} & \multicolumn{3}{c}{ResNet-50} \\ \cmidrule{2-13}
& $N_\mathrm{param}$ (MB) & TA (\%) & RA (\%) & $N_\mathrm{param}$ (MB) & TA (\%) & RA (\%) &$N_\mathrm{param}$ (MB) & TA (\%) & RA (\%) &$N_\mathrm{param}$ (MB) & TA (\%) & RA (\%)  \\ \midrule

Standard (32 bit) & $58.8$ & $92.59$ & $65.44$ & $58.8$ & $71.44$ & $36.76$ & $59.2$ & $61.28$ & $23.58$ & $95.7$ & $65.72$ & $29.65$  \\ 
Standard (16 bit) & $29.4$ \textcolor{blue}{($0.5 \times$)} & $88.05$ & $58.12$ & $29.4$ \textcolor{blue}{($0.5 \times$)} & $53.60$ & $21.17$ & $29.6$ \textcolor{blue}{($0.5 \times$)} & $51.66$ & $16.87$ & $47.9$ \textcolor{blue}{($0.5 \times$)} & $40.27$ & $20.26$  \\ 
\midrule

Data Augmentation (32 bit) & $58.8$ \textcolor{blue}{($1 \times$)} & $89.58$ & $84.34$ & $58.8$ \textcolor{blue}{($1 \times$)} & $67.34$ & $56.95$ & $59.2$ \textcolor{blue}{($1 \times$)} & $52.11$ & $43.64$ & $95.7$ \textcolor{blue}{($1 \times$)} & $59.17$ & $47.96$  \\
Data Augmentation (16 bit) & $29.4$ \textcolor{blue}{($0.5 \times$)} & $83.18$ & $74.93$ & $29.4$ \textcolor{blue}{($0.5 \times$)} & $60.58$ & $51.19$ & $29.6$ \textcolor{blue}{($0.5 \times$)} & $46.66$ & $35.01$ & $47.9$ \textcolor{blue}{($0.5 \times$)} & $39.19$ & $26.03$ \\
\midrule

All Models (32 bit) & $1177.6$ \textcolor{blue}{($20 \times$)} & $92.59$ & $88.97$ & $1177.6$ \textcolor{blue}{($20 \times$)} & $71.44$ & $64.72$ & $1184.0$ \textcolor{blue}{($20 \times$)} & $61.28$ & $51.55$ & $1913.6$ \textcolor{blue}{($20 \times$)} & $65.72$ & $55.97$ \\ 
All Models (16 bit) & $588.8$ \textcolor{blue}{($10 \times$)} & $88.05$ & $82.05$ & $588.8$ \textcolor{blue}{($10 \times$)} & $53.60$ & $52.96$ & $592.0$ \textcolor{blue}{($10 \times$)} & $51.66$ & $36.72$ & $956.8$ \textcolor{blue}{($10 \times$)} & $40.27$ & $23.00$  \\
\midrule
Standard+$\textbf{RWS}_\mathrm{Full}$ & $1177.6$ \textcolor{blue}{($20 \times$)} & $92.59$ & $75.35$ & $1177.6$ \textcolor{blue}{($20 \times$)} & $71.44$ & $52.58$ & $1184.0$ \textcolor{blue}{($20 \times$)} & $61.28$ & $43.63 $ & $1913.6$ \textcolor{blue}{($20 \times$)} & $65.72$ & $53.64$  \\
Standard+$\textbf{RWS}_\mathrm{Shallow}$ & $101.0$ \textcolor{blue}{($1.7 \times$)} & $92.59$ & $84.86$ & $101.0$ (\textcolor{blue}{$1.7 \times$)} & $71.44$ & $58.78$ & $101.4$ \textcolor{blue}{($1.7 \times$)} & $61.28$ & $44.66$ & $131.4$ \textcolor{blue}{($1.4 \times$)} & $65.72$ & $52.84$  \\
Standard+$\textbf{RWS}_\mathrm{Shallow, 16 bits}$ & $79.9$ (\textcolor{blue}{$1.4 \times$)} & $92.59$ & $84.76$ & $79.9$ \textcolor{blue}{($1.4 \times$)} & $71.44$ & $58.62$ & $80.3$ \textcolor{blue}{($1.4 \times$)} & $61.28$ & $44.25$ & $113.6$ \textcolor{blue}{($1.2 \times$)} & $65.72$ &$52.81$  \\
Standard+$\textbf{RWS}_\mathrm{Shallow, 8 bits}$ & $69.4$ \textcolor{blue}{($1.2 \times$)} & $92.59$ & $82.99$ & $69.4$ \textcolor{blue}{($1.2 \times$)} & $71.44$ & $53.52$ & $69.7$ \textcolor{blue}{($1.2 \times$)} & $61.28$ &$39.40$ & $104.7$ \textcolor{blue}{($1.1 \times$)} & $65.72$ &$52.79$ \\

 \bottomrule
\end{tabular}}
\end{table*}

\end{document}